\documentclass[runningheads]{llncs}

\usepackage[T1]{fontenc}
\usepackage{graphicx}
\usepackage{hyperref}
\usepackage{amsmath} 
\usepackage{color}

\definecolor{mygreen}{rgb}{0.0,0.5,0.6}
\definecolor{mygray}{rgb}{0.5,0.5,0.5}
\definecolor{mymauve}{rgb}{0.58,0,0.82}
\definecolor{cb_orange}{rgb}{1.0,0.51,0.0}
\definecolor{cb_blue}{rgb}{0.22,0.49,0.72}
\definecolor{cb_green}{rgb}{0.3,0.67,0.29}
\definecolor{cb_red}{rgb}{0.89,0.1,0.11}
\definecolor{cb_pink}{rgb}{1, 0, 0.4}

\begin{document}
\title{RLCorrector: Reinforced Proofreading for Cell-level Microscopy Image Segmentation}
\titlerunning{Reinforced Proofreading for Cell-level Microscopy Image Segmentation}
%
\author{
    Khoa Tuan Nguyen\inst{1}, Ganghee Jang\inst{1}, Tran Anh Tuan\inst{2}, and Won-ki Jeong\inst{1}
}
%
%
%
\authorrunning{Khoa et al.}
%
%
\institute{
    Department of Computer Science \& Engineering,
    Korea University
    \and 
    Max Planck Institute for Informatics, Saarland University\\
        \email{\{ntkhoa,jnggh,wkjeong\}@korea.ac.kr, atran@mpi-inf.mpg.de}
}
%
%
%
%
\maketitle              
\begin{abstract}
Segmentation of nanoscale electron microscopy (EM) images is crucial but still challenging in connectomics research. 
%
One reason for this is that none of the existing segmentation methods are error-free, so they require proofreading, which is typically implemented as an interactive, semi-automatic process via manual intervention. 
Herein, we propose a fully automatic proofreading method based on reinforcement learning that mimics  
the human decision process of detection, classification, and correction of segmentation errors. 
%
We systematically design the proposed system by combining multiple reinforcement learning agents in a hierarchical manner, where each agent focuses only on a specific task while preserving dependency between agents. 
Furthermore, we demonstrate that the episodic task setting of reinforcement learning can efficiently manage a combination of merge and split errors concurrently presented in the input. 
We demonstrate the efficacy of the proposed system by comparing it with conventional proofreading methods over various testing cases.

\end{abstract}

\section{Introduction}

%
%
Connectomics is a research field of investigating cellular-level neural connections in the brain~\cite{deweerdt2019map}.
Nanoscale electron microscopy (EM) images are typically used to resolve cell-level neuronal structures (e.g., dendritic spine necks and synapses) of only tens of nanometers in size~\cite{Helmstaedter2013}.  
Because the raw data size of EM serial sections of a small tissue sample can easily reach hundreds of terabytes, the need to develop high-throughput and automatic image processing algorithms has been growing in the past decade.

\begin{figure}[tp]
    \centering
    \includegraphics[width=0.8\textwidth]{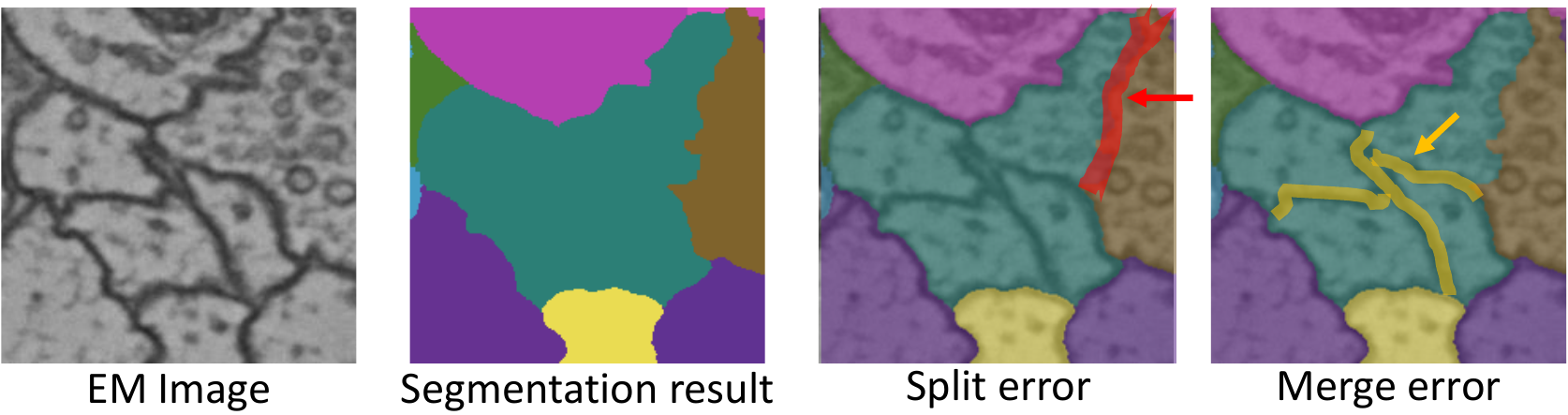}
    \caption{Example of segmentation errors for proofreading. Merge error occurs when more than two independent objects are incorrectly merged and assigned the same label. Split error occurs when single object is incorrectly split into multiple segments and assigned different labels.
    }
    \label{fig:errortype}
\end{figure}

%
%
Recent advances in deep learning have demonstrated the significant potential in the high-throughput, automatic segmentation of connectome images.
Many existing segmentation methods are based on pixel-level classification using convolutional neural networks (CNNs) and instance clustering~\cite{quan2016fusionnet,6122,Meirovitch_2019_CVPR}. 
However, such methods are not perfect and prone to errors, particularly when applied to real data, which requires manual proofreading by humans. 
Existing proofreading methods are primarily based on the interactive manual correction of either merge or split errors (see Figure~\ref{fig:errortype}) using an intuitive user interface and visualization~\cite{haehn2014design,knowles2013mojo,gonda_vice_2021}. 
Even with the support of such interactive tools, manual proofreading is a time-consuming and labor-intensive task, resulting in a bottleneck in the connectome analysis workflow.

To address the issues outlined above, fully automated proofreading approaches have been developed.
Zung \textit{et al.} \cite{zung_error_2017} designed an automatic proofreading algorithm at the level of the neuron reconstruction. 
The authors assumed that split errors are rare in the initial segmentation result, so they formulated the algorithm to prune wrongly merged super-voxels iteratively. 
Haehn \textit{et al.}~ \cite{haehn2018guided} designed a ranking system from a the CNN-based error detector for cell-level segmentation.
The above methods consist of multiple sub-tasks, such as detecting erroneous location, classifying error types, and correcting errors,  applied in a prioritized brute-force manner.
Moreover, there are many erroneous regions across the entire image, each of which requires application of different combination of above sub-tasks, making the proofreading process inefficient. 
%

%
%
To address these issues, we propose a novel automatic proofreading method based on reinforcement learning (RL), \texttt{RLCorrector}, for cell-level microscopy image segmentation.
The main motivation of this work stems from the following observation: Unlike general pixel-level image editing, our proofreading task can be regarded as an iterative decision-making process that consists of error location and error corrector selections. 
By designing the discrete action space and environment to model this decision process, the human proofreading process can be successfully mimicked by RL agents. 
To the best of our knowledge, this is the first RL-based proofreading system that operates fully automatically without human intervention, which can be a novel addition to the recent effort of using RL in various image processing problems~\cite{furuta_pixelrl_2020,Uzkent_2020_CVPR,araslanov2019actor,Tuan_2021_CVPR}. 
We show that our method outperforms conventional CNN-based methods in both error correction performance and execution time.

\section{Method}

\subsection{System Overview}

\begin{figure*}[tp]
    \centering
    \includegraphics[width=0.9\textwidth]{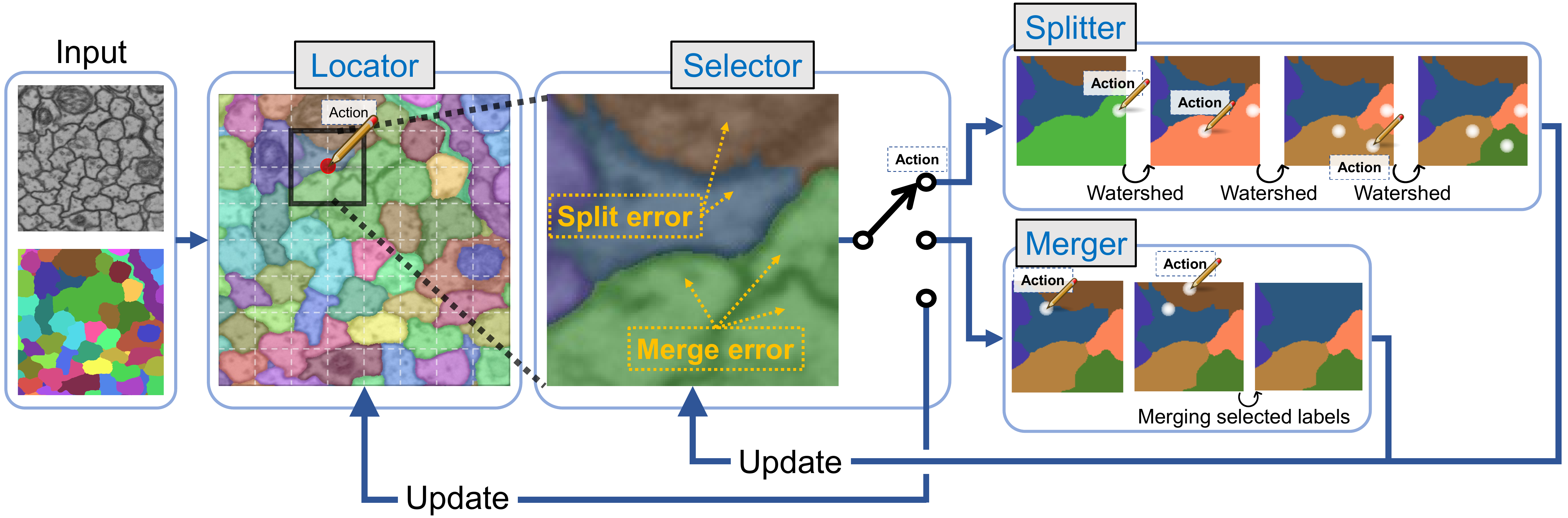} 
    \caption{Overview of the proposed RL proofreading system. The human decision process is mapped to the agent's actions. Updates are from individual correctors and used for decision making at the next time step.} 
    \label{fig:system_diagram}
\end{figure*}

The proofreading process of \texttt{RLCorrector} consists of identifying a patch containing errors (the locator agent), selecting an error correction method (the selector agent), and correcting errors (the merger and the splitter agents) as depicted in Figure~\ref{fig:system_diagram}. 
The locator's task is to identify an error location (i.e., a patch) on a coarse grid, and this identification is repeated until no more erroneous region are left. 
On the erroneous patch selected by the locator, the selector repeatedly chooses either the splitter or the merger to fix the error until no further error is found.
As for the reward metric, we employ the circuit reconstruction from
EM images (CREMI) score~\cite{cremi_dataset} from the MICCAI 2016 challenge. 
While evaluating the  performance using our test set, the environment  can also perform termination of an episode when there is no notable change on the label map over time.

\subsection{Details of RL Agents}
\label{subsubsec:rlagents}

\begin{figure}[tp]
    \centering
    \includegraphics[width=0.8\textwidth]{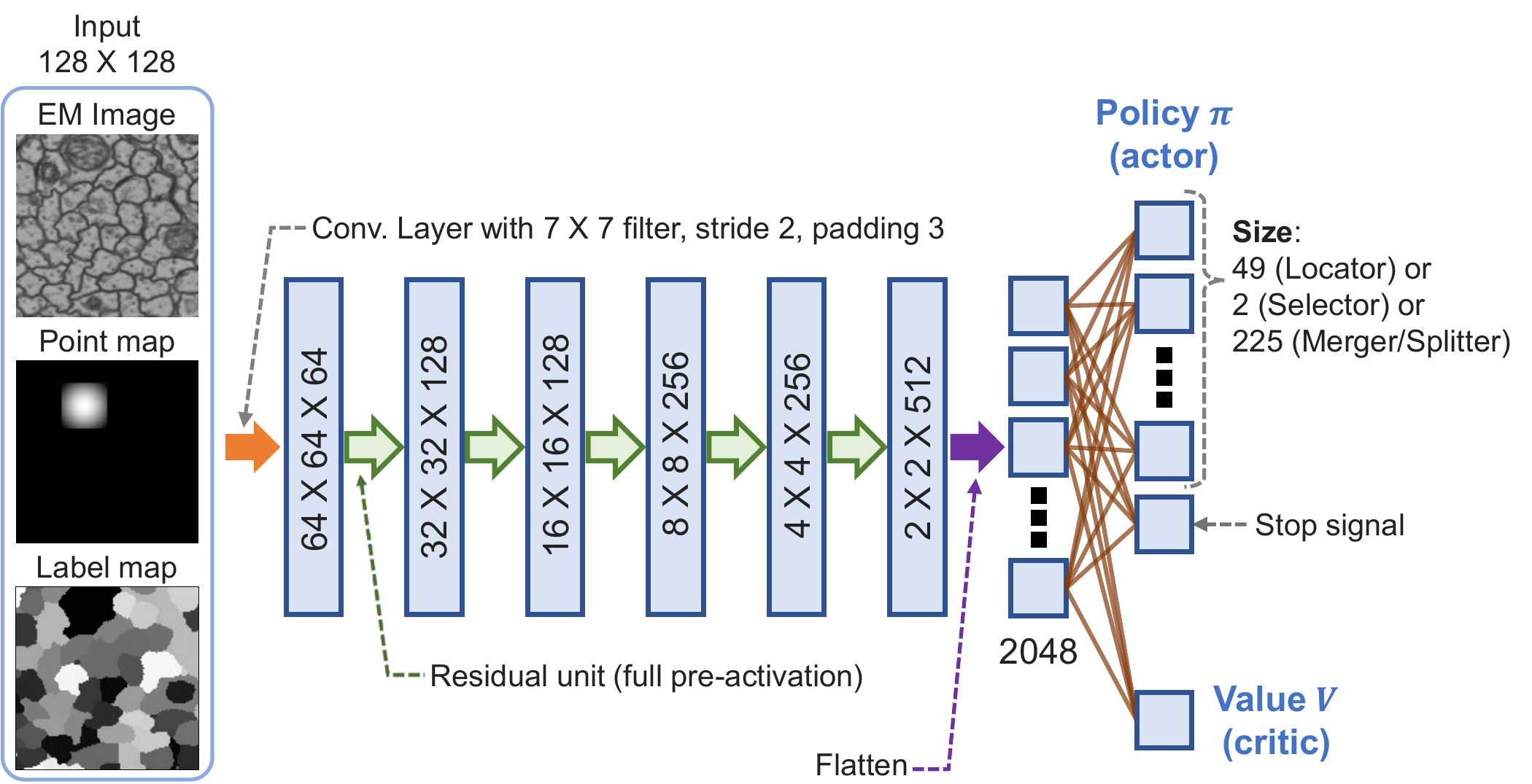}
    \caption{Neural network architecture of the RL agent. The point map is omitted when using the selector agent. Output vector size of actor varies with agents. } 
    \label{fig:agent}
\end{figure}

The architecture of the RL agent is based on the asynchronous advantage actor--critic method~\cite{mnih2016asynchronous} shown in Figure~\ref{fig:agent}. The input consists of three channels -- an EM image, a point map, and a label map. 
The point map keeps track of previous action points, and the label map shows the current segmentation result after applying the error correcting actions. 
The neural network for the agent 
consists of one convolutional layer (orange arrow) followed by five residual units (green arrows) of full pre-activation \cite{he2016identity} and a fully connected layer is added after flattening (violet arrow). 
The large blue box represents an output feature vector from a layer or residual unit, and the small blue box indicates a logit function. 
The output of the actor consists of many logits, whereas that of the critic has a single logit.
The actor's output vector size is the number of actions, which depends on each agent's action space, as shown in Figure \ref{fig:agent}.

\subsubsection{Locator.}
We used a 7 $\times$ 7 two-dimensional (2D) grid to define a finite action space with 49 locations, where the selected point collects four adjacent squares with dashed lines (see Figure~\ref{fig:system_diagram}). 
We generated the point map by applying a Gaussian kernel to provide a decent amount of field-of-view to the neural network.
As for the reward, we sought to evaluate the quality of merge and split error correction to encourage the selection of the patches with errors, and we measured the quality using a CREMI score.
When an action $c_{t}$ is performed at time step $t$, the selected patch $P_{c_t}$ and its label map $L_t$ are forwarded to the selector.
When $P_{c_t} \in E_{t}$ (the set of erroneous patches), a positive reward of 1 is received. 
$R_{diff}$ is added to promote the locator to select the patch with errors (i.e., the CREMI score of the resulting label map $L_{t+1}$ returned from the selector is reduced). 
Training the stop signal is important because avoiding wrong corrections on patches with no errors is critical to the overall performance. 
Thus, we give a high reward of 2 when it properly stops and a penalty of $-2$ when wrong corrections are made on the non-error area.
There are two proper termination conditions during training time: One is when there are no further CREMI score improvements, whereas the other is when the CREMI score is very low because of a lack of errors in the selected patch.
The entire reward function for the locator is described in Equation~\ref{eqn:R_locator}: 

\begin{equation}
    \label{eqn:R_locator}
    R_{t+1} = 
    \begin{cases}
        2,          &\text{if stop signal is correct,} \\
        1 + R_{diff}, &\text{if $CREMI(L_{t+1}) < CREMI(L_{t})$,} \\
                      &\text{      and $P_{c_t} \in E_t$,} \\
        -1,           &\text{if $CREMI(L_{t+1}) \geq CREMI(L_{t})$,} \\
                    &\text{      and $P_{c_t} \in E_t$,} \\
        -2,          &\text{if stop signal is wrong}
    \end{cases}
\end{equation}

\begin{equation}
    \label{eqn:$R_diff$}
    R_{diff} = \frac{CREMI(L_{t})-CREMI(L_{t+1})}{CREMI(L_{t})} 
\end{equation}

\subsubsection{Selector.}

The selector performs a series of selections among the merger, splitter, and stop actions. 
The agent launches a new episode to fix the corresponding errors of the selected agent.
The stop action ends the current episode made by the selector, and the execution is returned to the locator. 
The input vector size to the agent is 128 $\times$ 128 $\times$ 2, as there is no point map. 
We scaled down the values for stop rewards to 1 and -1 to make the selector focuses more on the decision of selecting a proper error corrector.
The positive reward is for $CREMI(L_{t+1}) < CREMI(L_{t})$, as no grid actions are involved. 
The whole reward function is shown in Equation \ref{eqn:R_selector}.
We set the maximum length of a selector episode to six to encourage exploration during training (four for the inference). 
See the supplementary for an example of a selector episode (Supp-Fig.~1).
Each label map under the selector time step is the result of a merger or splitter episode. 
At $t=1$, a wrong pair of segments is merged, but by applying another merger and a splitter episode, the errors are completely fixed. 
This demonstrates the manner in which the selector explores the action space and solves the problem.

\begin{equation}
    \label{eqn:R_selector}
    R_{t+1} = 
    \begin{cases}
        1 + R_{diff}, &\text{if $CREMI(L_{t+1}) < CREMI(L_{t})$,} \\
        1,          &\text{if stop signal is correct,} \\
        -1,           &\text{if $CREMI(L_{t+1}) \geq CREMI(L_{t})$,} \\
                    &\text{else if  stop signal is wrong}
    \end{cases}
\end{equation}

\subsubsection{Splitter.}

The splitter agent is similar to the locator except a finer grid space is employed (i.e., the total actions for the grid selection are 15 $\times$ 15). 
For a selected grid point, the watershed algorithm is applied to split the given segment into two.
The altitude map for the watershed processing is generated by Gaussian smoothing applied on the EM image. (Note that the cell membrane is darker than other regions, which makes a natural boundary for watershed). 
We set the maximum episode length to six. 
(See Supp-Fig.~2 for an example of a splitter episode). 
Note that the label map is updated with new action at the end of $t=2$ and $t=3$. 
The reward function of the splitter is similar to that of the locator except the stop signal:

\begin{equation}
    \label{eqn:R_splitter}
    R_{t+1} = \begin{cases}
    1 + R_{diff}, &\text{if $CREMI(L_{t+1}) < CREMI(L_{t})$,} \\
                &\text{and $P_{c_t} \in E_t$,} \\
                1, &\text{if stop signal is correct,} \\
                -1, &\text{if $CREMI(L_{t+1}) \geq CREMI(L_{t})$,} \\
                &\text{      and $P_{c_t} \in E_t$,} \text{else if  stop signal is wrong} 
                \end{cases}
\end{equation}

\subsubsection{Merger.}
The merger's reward function is almost identical to that of the splitter, but the episode formulation of the merger is different because the merging operation requires at least one pair of neighboring segments. 
Each action selects a grid point, and a labeled segment by the selected point will be a part of a set of segments to be merged.
We apply the merging operation once every two time steps (i.e., after selecting two grid points). 
An example of a single merge operation after two time steps of a merger episode is shown in the supplementary data (Supp-Fig.~3). 
A negative reward is given to prevent an incorrect merge that increases the CREMI score. 

\section{Results}
\label{sec:results}
\subsection{Experiment setup}

We compared our scheme with the method by Haehn~\textit{et al.}~\cite{haehn2018guided} with the CREMI dataset~\cite{cremi_dataset} which consists of three sets -- A, B, and C -- each of which comprises 125 slices of 1250 $\times$ 1250 in size.
The first 92 slices were used for training, 23 for validating, and 10 for testing.
Training was performed using the bottom-up approach. 
The merger and splitter were trained first, followed by the selector, and then the locator was trained at the end. 
For training the agents, we used the Adam optimizer and set the learning rate to $10^{-5}$. 
We used the asynchronous advantage actor--critic method~\cite{mnih2016asynchronous} based on the source code from \cite{A3G}.

For a fair comparison with Haehn~\textit{et al.}~\cite{haehn2018guided}, we shared our agent's backbone network with Haehn's proofreading classifier to give a similar model capacity. 
In addition, we set the input image size to be the same as that of our splitter and selector. 
For fully automated proofreading on a single image, we applied a sliding window method to Haehn's proofreading algorithm.

\subsection{Performance comparisons}
\label{subsec:performancecomparisons}

\subsubsection{Patch-level Performance.}
The first experiment involved evaluating patch-level performance of the agents. 
We set the patch size as $128 \times 128$, and this was cropped centered at the randomly chosen grid-action point in the image level grid ($512 \times 512$). 
Table~\ref{tab:perpatch} shows the per-patch error correction performance measured on 1000 patches.
We used three synthetically generated test sets with different error types -- merge error only, split error only, and both errors. 
In our method, ``Static'' is the version without the selector, so that the merger and splitter are applied once in a pre-defined order, as in Haehn's. 
The ``Selector'' is the full version of \texttt{RLCorrector} with a dynamic selection of error correctors by the selector agent. 
As shown in Table~\ref{tab:perpatch}, our method significantly improves the error correction performance, even without the selector. (However, Haehn's merger has an edge over ours with a tiny margin for the ``split error only`` case). 
It should also be noted that our splitter (i.e., merge error corrector) is much stronger than Haehn's. 
The performance of our method is improved further when the selector kicks in, which makes our method outperform Haehn's method in all cases.

\begin{table}[htb]
    \centering
    \caption{Per-patch error correction performance evaluation in the average CREMI score (lower is better). The best result for each error type is marked in bold.}
    \begin{tabular}{| p{3cm} ||  p{2cm} ||  p{2cm} || p{2cm} | p{2cm} | }
        \hline
                         &  Test set  & Haehn's \cite{haehn2018guided} & \multicolumn{2}{c|}{Ours}   \\
        \cline{4-5}      &            &         &  Static & Selector\\
        \hline
        \hline
        Merge error only & 0.173     & 0.081   &  0.032  & \textbf{0.028}\\ 
        \hline
        Split error only & 0.152     & 0.019   &  0.020  & \textbf{0.014}\\ 
        \hline
        Combined         & 0.272     & 0.111   &  0.045  & \textbf{0.040} \\ 
        \hline
    \end{tabular}
    \label{tab:perpatch}
\end{table}

\subsubsection{Image-level Performance.}
In this experiment, we assessed the performance at the image level with real segmentation errors. 
For this, we used the attention U-Net~\cite{oktay2018attention} to generate the initial segmentation result on the CREMI dataset. 
Since the locator action space is 7 $\times$ 7 for a patch size (i.e., the input size of the error corrector) of 128 $\times$ 128 with a stride of 64, the input image is split into sub-images of size 512 $\times$ 512 with a stride of 256. 
In each sub-image, we applied a sliding-window scheme to fix errors except for our \texttt{Locator-Selector} because it can directly select the location of the erroneous patch.
As shown in Table~\ref{tab:unetsegmentation}, all three variations of our method outperformed Haehn's methods in terms of both error conrrection performance and speed. 
Especially, our method achieved a lower CREMI score by 26.5\% on average in set A, while 54.6\% in set B, and 50.7\% in set C over Haehn's method. 
Execution times were reduced by 95.7\%, 94.7\%, and 94.6\% for our  \texttt{Locator-Selector, Sliding-Static, and Sliding-Selector} methods, respectively . 
It should be noted that the selector contributes to improving error correction performance (low CREMI score), whereas the locator contributes to reducing the execution time over the static method, as we expected.
\begin{table}[tp]
    \centering
    \caption{Image-level proofreading performance on U-Net segmentation results measured in average CREMI score (lower is better) and execution time.}
    \begin{tabular}{| c | c|| p{1.5cm} || p{1.6cm} || p{1.3cm} | p{1.3cm}  | p{1.3cm} |}
        \hline
        \multicolumn{2}{| c ||}{} & Test set & Haehn's \cite{haehn2018guided} & \multicolumn{3}{c|}{Ours}  \\
        \cline{5-7} 
        \multicolumn{2}{| c ||}{}  & & Sliding & Sliding & Sliding & Locator \\
        \multicolumn{2}{| c ||}{}  & & & (Static) & (Selector) & (Selector) \\
        \hline
        \hline
        CREMI A & Ave. score    & 0.481 & 0.312 & 0.228     & \textbf{0.219}    & 0.241 \\
        \cline{2-7}
        & Ex. time (min.)      & N/A    & 324.5 &  19.3    &  18.6             & \textbf{14.9}  \\
        \hline

        \hline
        CREMI B & Ave. score    & 1.429 & 0.535 & 0.270     & 0.250             & \textbf{0.208}  \\
        \cline{2-7}
        & Ex. time (min.)      & N/A    & 358.3 & 18.8      & 18.4              & \textbf{14.7}  \\
        \hline
        
        \hline
        CREMI C & Ave. score    & 0.967 & 0.706 & 0.358     & \textbf{0.337}    & 0.349  \\
        \cline{2-7}
        & Ex. time (min.)      & N/A    & 388.9 & 19.7      & 19.7              & \textbf{16.7}  \\
        \hline
    \end{tabular}
    \label{tab:unetsegmentation}
\end{table}

\begin{figure}[htp]
    \centering
    \includegraphics[width=0.9\textwidth]{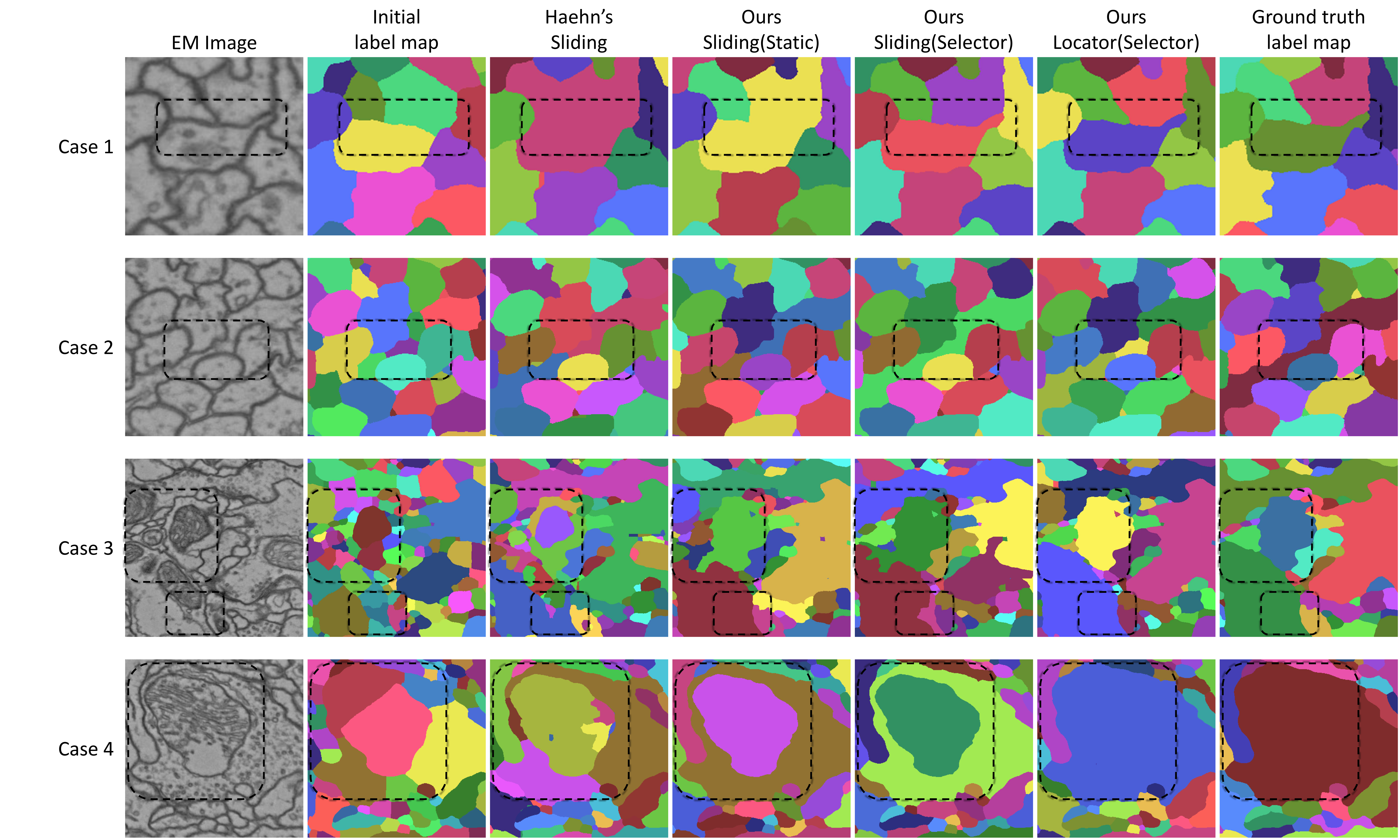}
    \caption{Comparisons of error corrector's behavior. Case 1 is when both the selector and locator worked as intended. In case 2, the locator fails to spot the error location, even though the selector can handle it. Cases 3 and 4 show that the locator contributed performance improvement even when sliding methods fail.
 } 
    \label{fig:result_real_seg}
\end{figure}

\subsubsection{Visual Assessment.}
To have a better understanding of our method, we visually inspected how our selector and locator contribute to performance improvement. 
Figure~\ref{fig:result_real_seg} compares several example cases in which the error correctors behaved differently; these cases are described as follows:
\begin{itemize}
\item \textbf{Case 1:} The locator spotted the erroneous areas correctly. 
In this area, the selector could choose agents adequately to fix the errors, whereas static methods (Haehn's and our static method) failed. 
A more detailed step-by-step description can be found in the supplementary (see Supp-Fig.~4). 
\item \textbf{Case 2:} The locator missed the error location so the error was not fixed, but the selector was able to fix the error.
\item \textbf{Case 3:} Some errors were newly created during the process (because of false error correction), but the sliding window methods could not fix them. However, selective choice of erroneous areas in proper order by the locator can avoid this issue.
\item \textbf{Case 4:} Another case in the CREMI C set when errors could not be fixed without help from the locator.
\end{itemize}

\section{Conclusion and Future Work}

Inspired by the human decision-making process, we introduced a novel, fully automatic proofreading method based on reinforcement learning for cell-level microscopy image segmentation. 
In this work, we modeled each task in the proofreading process using a reinforcement learning agent and hierarchically combined them to design a multi-agent system. 
%
%
We demonstrated that the dynamic nature of our system significantly improved segmentation performance while reducing execution time compared with conventional proofreading methods. 
%
%

Despite the performance benefit of the proposed method, there is still room for improvement. 
Because of the coarse and discrete nature of the action space, handling small fragments is difficult. 
We plan to address this problem by employing a continuous action space. 
Furthermore, extension to three dimensions is another interesting future research direction.

\subsubsection{Acknowledgements.} 

TBD


%
%
%
\bibliographystyle{splncs04}
\bibliography{ref}

\end{document}


\title{Supplementary}
%
%
\author{Submission ID: 1892}
%
%
%
\authorrunning{Anonymous}
%
%
\institute{Anonymous Organization}
%
%
%
%
\maketitle              
%
%


\begin{figure}[htp]
    \centering
    \includegraphics[width=1\textwidth]{figs/Selector_khoa.pdf}
    \caption{Example of selector episode. Label map of each time step is from correctors.} 
    \label{fig:selector}
\end{figure}

\begin{figure}[htp]
	\centering
	\includegraphics[width=1\textwidth]{figs/Splitter_khoa.pdf}
	\caption{Example of splitter episode. Environment runs watershed algorithm with given actions as valleys.}
	\label{fig:splitter}
\end{figure}

\begin{figure}[htp]
    \centering
    \includegraphics[width=0.9\textwidth]{figs/Merger_khoa.pdf}
    \caption{Example of merger episode. Two time steps are required for a merge operation.}
    \label{fig:merger}
\end{figure}


\begin{figure}[htp]
    \centering
    \includegraphics[width=0.9\linewidth]{figs/example_fixing.pdf}
    \caption{Some sliding window steps of \textbf{Case 1}. 
    Haehn's sliding and our sliding(static) added merge error, but sliding(selector) fixed error by avoiding applying merger. Last row is to show that patch location is pointed by locator. } 
    \label{fig:example_fixing}
\end{figure}

%
%
%
\bibliographystyle{splncs04}